%% file: main.tex
\documentclass[10pt,twocolumn,letterpaper]{article}

\usepackage{cvpr}      % To produce the REVIEW version
\usepackage{amsmath, graphicx, bm, caption, subcaption} %subfig}
\input{preamble}

\definecolor{cvprblue}{rgb}{0.21,0.49,0.74}
\usepackage[pagebackref,breaklinks,colorlinks,citecolor=cvprblue]{hyperref}

\def\confName{CVPR}
\def\confYear{2024}

\title{Semantically Grounded QFormer for Efficient Vision Language Understanding}

\author{Moulik Choraria$^*$\\
University of Illinois at Urbana Champaign\\
{\tt\small moulikc2@illinois.edu}
\and
Xinbo Wu$^*$\\
University of Illinois at Urbana Champaign\\
{\tt\small xinbowu2@illinois.edu}
\and
Sourya Basu\\
University of Illinois at Urbana Champaign\\
{\tt\small sourya@illinois.edu}
\and
Nitesh Sekhar\\
Amazon\\
{\tt\small seknites@amazon.com}
\and
Yue Wu\\
Amazon\\
{\tt\small wuayue@amazon.com}
\and
Xu Zhang\\
Amazon\\
{\tt\small xzhnamz@amazon.com}
\and
Prateek Singhal\\
Amazon\\
{\tt\small prtksngh@amazon.com}
\and
Lav R. Varshney\\
University of Illinois at Urbana Champaign\\
{\tt\small varshney@illinois.edu}
}

\begin{document}
\maketitle
\def\thefootnote{*}\footnotetext{These authors contributed equally to this work}\def\thefootnote{\arabic{footnote}}
% \begin{abstract}
    
% \end{abstract}
% %\input{sec/0_abstract}    
% \input{sec/1_intro}
% \input{sec/2_formatting}
% \input{sec/3_finalcopy}

\begin{abstract}
   General purpose Vision Language Models (VLMs) have received tremendous interest in recent years, owing to their ability to learn rich vision-language correlations as well as their broad zero-shot competencies. One immensely popular line of work utilizes frozen unimodal models, by bridging vision representations to language using a trainable module called the QFormer. However, this method relies heavily on large-scale multimodal pretraining with huge computational overheads. To that end, we propose a more efficient framework for QFormer-based vision-language alignment. Our key idea relies on the observation that QFormer latents correspond more strongly to the frozen LLM's intermediate latent space. Consequently, instead of using QFormer latents as inputs to the LLM, we alter the framework by using the latents to directly condition the LLM latent space for image-to-text generation. We demonstrate the effectiveness of our approach against existing baselines in improving the efficiency of vision-language pretraining.
\end{abstract}

%%%%%%%%% BODY TEXT
\section{Introduction}
\label{sec:intro}

Vision-language models (VLMs) are a rapidly emerging topic of research due to their ability to harness multi-modal inputs for simultaneous visual and textual comprehension for diverse applications. Following the success of general purpose large language models \cite{TomBMRSKA2020, HyungCW2022}, efforts are underway towards analogous general purpose VLMs with few-shot learning capabilities. As such, methods relying on instruction finetuning \cite{JasonWBZGYLDDL2022, JasonWSBIXCLZ2023} are appealing due to their ability to exploit natural language understanding; one particular line of work that combines it with multi-modal fusion of frozen language and vision models has received tremendous attention \cite{JunnanLLSH2023, WenliangDLLTZWLFH2023}. Specifically, \cite{JunnanLLSH2023} relies on combining frozen unimodal representations via a multi-stage pretrained QFormer, followed by multi-task instruction finetuning, yielding state-of-the-art zero-shot performance on a variety of tasks \cite{WenliangDLLTZWLFH2023}. Following their success, a slew of recent methods explore alternate strategies, while building on top of the QFormer base \cite{ZijiaZGYCSZYL2023, KunChangLHWLWLWWQ2023}. \\

The success of \cite{JunnanLLSH2023} heavily relies on the large-scale two-stage pretraining on the QFormer. In the first stage, the QFormer (initialized as a BERT model) is trained by a combination of self-supervised losses to learn a joint representation space (called Queries). % which include image-text contrastive learning, image-text matching, and image to text generation. 
In the next pretraining stage, the output Queries of the QFormer are fed to the frozen LLM to act as conditioning for end-to-end image-to-text generation. Ablation studies  demonstrate that the two-stage large-scale pretraining plays a critical role in empowering the QFormer. 
%In the first stage, copious amounts of paired image-text multi-modal data is leveraged for training the QFormer to learn a joint representation space, by a combination of self-supervised losses, which include image-text contrastive learning, image-text matching, and image to text generation. In the next stage, the QFormer is used to condition the frozen LLM for end-to-end image to text generation. 

Despite the undeniable benefits however, the amount of data and compute required for pretraining at scale presents a very expensive hurdle, and remains inaccessible to the larger research community \cite{EmmaSGM2019, JulianTY2023}. In addition, curating diverse and unbiased multi-modal datasets requires substantial efforts to ensure VLMs do not inadvertently propagate harmful biases in the training data \cite{YiZWS2022}. Hence, there is a pressing need for more efficient strategies for training VLMs and to that end, we seek to improve the pipeline in \cite{JunnanLLSH2023, WenliangDLLTZWLFH2023}. 

Our contributions are as follows: we first analyze the QFormer representations, and its ease of modelling inputs for LLM conditioning. Specifically, we show that the QFormer representations are better suited to model the frozen LLM's intermediate representations. \textbf{Besides, we study existing alignment between different layers of language models and vision models.} Based on these insights, we propose an efficient alternative pipeline for QFormer-based vision-language modelling. Under controlled settings, we then demonstrate the effectiveness of our approach in both speeding up vision-language representation learning as well as improving performance.

\section{Proposed Method}

\subsection{Problem Formulation}
%This section describes our proposed pipeline, starting with a refresher on the QFormer, the key component for the BLIP models \cite{JunnanLLSH2023, WenliangDLLTZWLFH2023}, which is the primary focus of our work.
In this paper, we focus on the problem of training a general-propose VLM, from the lens of instruction tuning. Specifically, given an image $i$ and a text instruction/prompt $p$, the model should output an appropriate response corresponding to the inputs. Some example tasks include visual question answering where $p$ is a question based on the image, or image captioning where $p$ refers to the instruction of captioning the image. Within vision-language modelling, we focus on the QFormer-based line of work, that bridges frozen unimodal models via a smaller model (the QFormer \cite{JunnanLLXH2022, JunnanLLSH2023, WenliangDLLTZWLFH2023}) for effective vision-language learning.

\subsection{Revisiting InstructBLIP} 
InstructBLIP is one of the pioneering works on vision-language instruction tuning, utilizing the QFormer approach as a trainable bridge between frozen vision and language models. We first consider LLMs with encoder-decoder architectures \cite{JacobDCLT2019, HyungCW2022}. Then, the full InstructBLIP framework can be interpreted as four different modules. \\
\textbf{Image Encoder} $\{v(\cdot)\}$: This module extracts visual features from the image input $i$, using a frozen pretrained vision model.\\
\textbf{QFormer} $\{Q(\cdot, \cdot, \cdot)\}$: The QFormer takes in a set of learnable query tokens ($\bm{t}_q$), the prompt $p$, and the encoder vision features $v(i)$ and outputs visually informed query tokens~$\bm{t}_{qv}$. \\
\textbf{LLM (Encoder-Decoder)} $\{l_e([\cdot, \cdot])$, $l_d(\cdot)\}$: The encoder takes the concatenation ($[\cdot, \cdot]$ denotes concatenation) of the QFormer output tokens~$\bm{t}_{qv} = Q(\bm{t}_q, v(i), p)$ and the language prompt $p$ as input and outputs a latent encoding, which is then used by the decoder to generate the desired text output. Concisely, the InstructBLIP output is represented as:
\begin{equation}
    \omega_{insB} = l_d(l_e([ \bm{t}_{qv}, p])) = l_d(l_e([Q(\bm{t}_q, v(i), p), p])).
    \label{eq:1}
\end{equation}
\textbf{LLM (Decoder-only)} If instead of an encoder-decoder model, we consider a decoder-only model (add cite), we can treat the encoder to be an identity function i.e. $l_e(\cdot, \cdot) = (\cdot, \cdot)$ and the output representation becomes:
\begin{equation}
    \omega_{insB} = l_d(l_e([p, \bm{t}_{qv}])) = l_d([p, Q(\bm{t}_q, v(i), p)]).
    \label{eq:2}
\end{equation}
A key strength of the BLIP-2/InstructBLIP framework is that the image encoder~$v(\cdot)$, the LLM encoder $l_e(\cdot)$ (if present) and the LLM decoder $l_d(\cdot)$ are frozen during training, with only the query tokens and the QFormer being trainable. Within the QFormer, the query tokens are allowed to interact with the prompt text tokens as well as each other via self-attention layers, and with the image features via alternate cross-attention blocks. The goal is to output a set of visually informed query tokens that can elicit an appropriate response from the downstream LLM encoder and decoder. 

The QFormer~\cite{JunnanLLSH2023} requires two heavy pretraining stages: Stage 1 comprises QFormer representation learning via image-text matching (ITM), image-text contrastive learning (ITC), and image-to-QFormer text generation (ITG) using the image encoder. Herein, the goal of the ITC and ITM is to align text features and query tokens in the QFormer latent space, whereas ITG trains the QFormer language modeling head for text generation by training the query tokens to extract visual features. Then, in Stage 2, these QFormer representations (``queries'') are projected onto the input space of the frozen LLM using a trainable fully-connected layer. These inputs are used to condition the frozen LLM for end-to-end image to text generation. 

It is important to note that while in theory, Stage 1 pretraining requires more epochs, it only utilizes the frozen vision encoder and the QFormer model and is therefore a) requires less memory and b) LLM agnostic, since the same QFormer can be used as a plug-and-play model for different frozen LLMs, when initializing for Stage 2. On the other hand, pretraining in Stage 2 is LLM specific, which means adaptation to newer language models comes with additional computational burdens. Thus, we want to find a way to make this adaptation more efficient.

\begin{figure}
     \centering 
         \includegraphics[width=\linewidth]{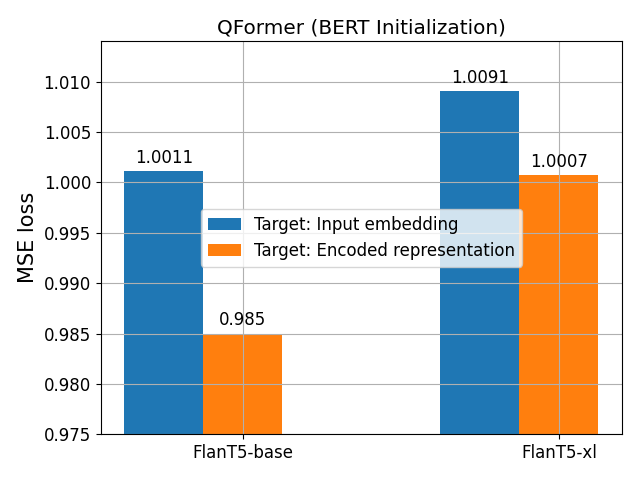}
         \caption{Experiment to study ease of modeling input embeddings vs encoder representations, the lower error indicates that the QFormer (BERT init) finds it easier to model the latter.}
        \label{fig:flan_bert}
\end{figure}

\begin{figure*}[!tbp]
  \centering
  \begin{minipage}[b]{0.528\textwidth}
\includegraphics[width=\textwidth]{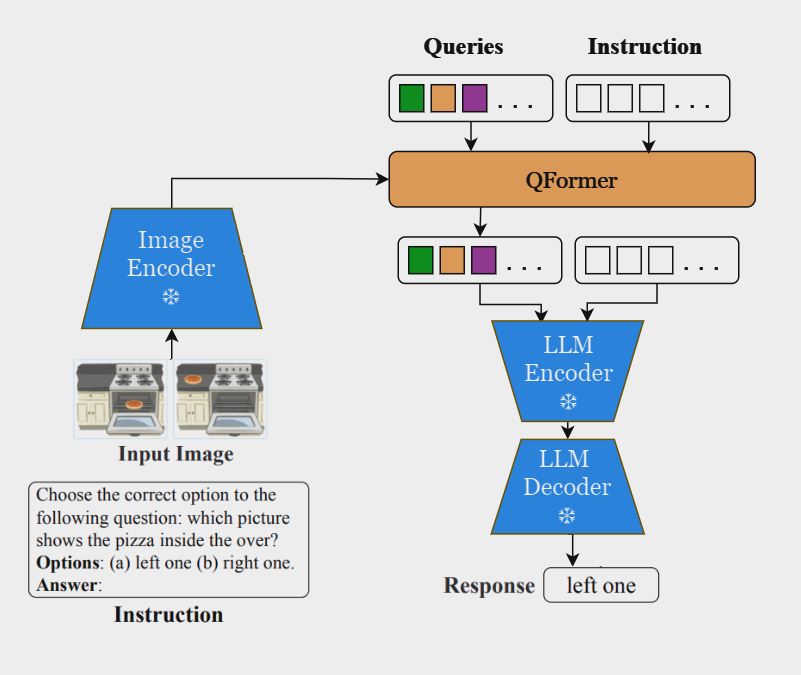}
    \subcaption{Standard QFormer}
  \end{minipage}
  \hfill
  \begin{minipage}[b]{0.45\textwidth}
\includegraphics[width=\textwidth]{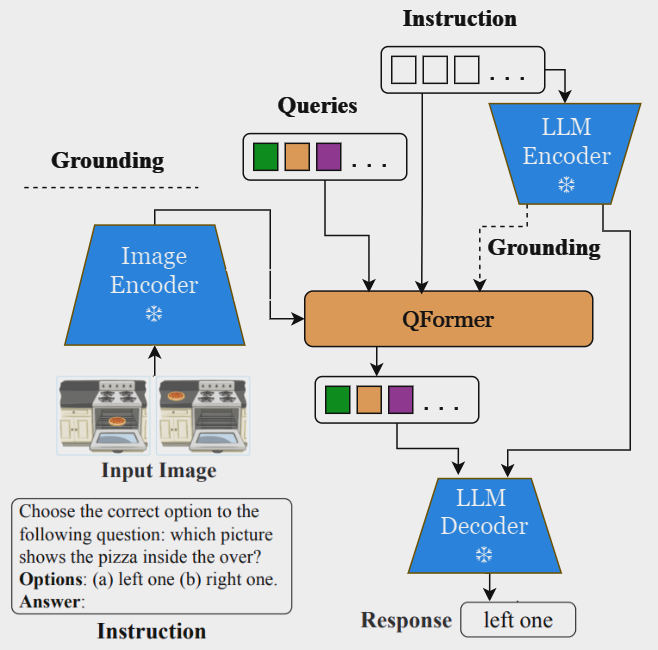}
    \subcaption{Grounded QFormer (Ours)}
  \end{minipage}
  \caption{Illustration of our framework for vision language fusion with the language grounded QFormer (right), distinguishing it from the standard design (left) as proposed in InstructBLIP \cite{WenliangDLLTZWLFH2023}.}
  \label{fig:proposal}
\end{figure*}

\subsection{Analyzing QFormer Representations}

Our first key observation stems from noting that the QFormer model is initialized as a language model, specifically a BERT encoder \cite{JacobDCLT2019}. Thus while the outputs of the QFormer (the encodings) can be used to model language using language modelling heads, these encodings themselves are imbued with semantic meaning. And this holds not just for BERT encoders, but also for general encoder language models \cite{JianmoNACMHCY2021}, wherein the encoder embeddings are often used for Sentence-level downstream tasks. However, note that in Stage 2, the QFormer latents are projected to the LLM input embedding space to condition vision-based generation. Thus, in essence, the standard QFormer tries to model the right syntactics for the frozen LLM, using representations based on the semantics of language. 

It is accepted wisdom that the semantic essence of the data often lies on a lower-dimensional manifold, embedded in a higher-dimension (syntactic) input space \cite{IanGBC2016}. For our application, we hypothesize that in terms of distributional similarity, the semantic latents of the QFormer correspond more strongly with the corresponding semantics of the LLM in the intermediate representations, as opposed to the LLM input text embeddings. Then, if the hypothesis holds, we could use the QFormer latents to directly model the LLM intermediates, to avoid this semantic-syntactic disparity in the standard QFormer pipeline and improve the learning efficiency of the framework.
% Therefore, it is often more efficient to model distributions in a lower-dimensional space \cite{DiederikKW2022, IanGAMXFOCB2014}.

To verify our hypothesis, we devise a simple experiment to compare the ease of modelling the LLM intermediates to modelling the input embeddings, using the QFormer outputs. For this, we consider a representative subset of the Wikitext-103 dataset \cite{StephenMXBS2016}. We consider the QFormer with default BERT initialization as in \cite{JunnanLLSH2023, WenliangDLLTZWLFH2023}, and for the LLM, two variants of the FlanT5 \cite{HyungCW2022} family of encoder-decoder LLMs, flanT5-base (220M) and flanT5-xl (3B). The choice is for two reasons: a) ease of LLM intermediate extraction, since we know the encoder outputs are semantically meaningful and b) it is one of two family of models considered in the InstructBLIP work \cite{WenliangDLLTZWLFH2023}. The experiment corresponds to a regression task; specifically, the same block of text is passed through the QFormer and the LLM. Then the outputs of the QFormer are projected via a linear projection (as in the standard framework) to match either the LLM input embeddings or the LLM encoder outputs, for the same text block via an L2 loss objective. For a fair comparison, we normalize both target distributions, input embeddings and encoder outputs, before training. The regression errors are reported in Fig. \ref{fig:flan_bert}, which demonstrate that for both models, there is a noticeable drop in error when modelling the encoder representations, supporting our hypothesis.

\subsection{Existing Alignment across Modalities}\label{sec:alignment}
Additionally, the Qformer is designed to align representations across both language and vision modalities. ~\citet{huh2024platonic} demonstrate that as vision models and language models scale up, their representational spaces converge across data modalities. Given that current deep learning models rely on hierarchical processing across layers, this motivates us to ask a finer-grained question: Does the alignments across modalities vary across layers?

To explore this, we measure the existing alignment between the representations of a pre-trained language model, flanT5-base ~\cite{HyungCW2022}, and a pre-trained vision model, the CLIP-based vision transformer Eva-clip-g/14 ~\cite{QuanSFWWC2023}, using the mutual KNN alignment metric proposed by ~\citet{huh2024platonic}. Since flanT5-base employs an encoder-decoder architecture, our experiments focus on its encoder, emphasizing its representation learning capabilities.

\begin{figure}
     \centering 
         \includegraphics[width=\linewidth]{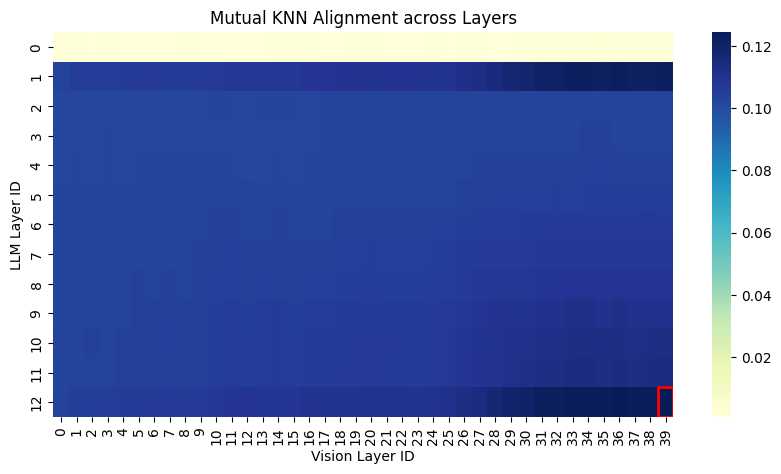}
         \caption{Mutual KNN alignment scores across the layers of an LLM (flanT5-base) and a vision transformer (Eva-clip-g/14) are presented as a heat map. The x-axis represents the layer IDs of the vision transformer, while the y-axis represents the layer IDs of the LLM. The maximum score is highlighted with a red box.
         }
        \label{fig:alignment_heatmap}
\end{figure}

For our measurements, we use the final token representation of the language model and the first token representation of the vision transformer, as these tokens are designed to aggregate the information from their respective inputs. Experiments were conducted on samples from COCO captioning validation set ~\cite{XinleiCFLVGDZ2015}, with K=10 neighbors, which has been shown to perform well in ~\citet{huh2024platonic}. The computed alignment scores are averaged across samples for each layer, and the results are visualized in Figure ~\ref{fig:alignment_heatmap}.

The heat map reveals that higher alignment scores are achieved for deeper layers of both models, with the highest scores observed at their final layers. This may suggest that the common practice of aligning representations from a deeper layer of a vision model to the initial layer of a language model ~\cite{JunnanLLSH2023} might be suboptimal due to low alignment scores. Instead, aligning deeper layers with higher alignment scores could yield better results, as corroborated by our follow-up experiments in Section ~\ref{sec:experiment}. This finding highlights the potential of using alignment to guide architecture design and further supports our hypothesis that aligning intermediate layers across modalities could enhance efficiency.

\subsection{Grounding the QFormer} 
The observation in the previous sections is key to our method, in that, instead of directly feeding the LLM, the QFormer output tokens directly model a visually informed encoder latent representation, which is then fed to the LLM decoder for text image-to-text generation. To aid this process, we propose grounding these representations by augmenting the QFormer queries with representations of the input prompts from the LLM encoder. These representations act as a reference/feedback for modeling the encoder latent representation efficiently. Furthermore, we integrate the Qformer output directly into the LLM decoder, bypassing its encoder, due to the better cross-modal alignment demonstrated in Section ~\ref{sec:alignment}.   
%drawing inspiration from classical notions of language grounding in cognitive science \cite{DebRP2002, DebR2002}

% More precisely, consider an image-prompt pair $(i, p) \in (I, T)$, the joint space of images and text, as the input to the VLM. Denote the frozen LLM model as $\ell(\cdot, [\cdot]) = d(e(\cdot, [\cdot]), [\cdot]): (T, [R]) \rightarrow T$, where the model takes in text inputs and [optionally] conditioning representations to generate text outputs, with $e(\cdot)$ and $d(\cdot)$ representing the encoder and decoder respectively. Next, the frozen vision model denoted as $v(\cdot): I \rightarrow I^*$, takes in images and extracts visual features in space $I^*$. Finally, the QFormer as $Q(\cdot, \cdot):(I^*, T) \rightarrow R$, takes in text inputs and frozen vision features and outputs a joint representation (queries) in space $R$, which is used to condition the LLM. Then the respective outputs, $\bm{\omega}_q$ for QFormer, and $\bm{\omega}_{gq}$ for our Grounded QFormer, may be denoted as:
We now provide a more precise characterization of our method, focusing on encoder-decoder LLMs. First,  consider $[\phi, p] = p$ as the concatenation of the prompt with an empty string, which serves as the input for the LLM encoder.  Next, denote the language grounded QFormer as $Q_g(\cdot, \cdot, \cdot)$ which accepts as input, language augmented queries ($[\bm{t}_q, l_e([\phi, p])]$), the image representations $v(i)$ and the prompt $p$, to output grounded queries $\bm{t}_{qv-g} = Q_g([\bm{t}_q, l_e([\phi, p])], v(i), p)$. Then, InstructBLIP framework with the grounded QFormer ($Q_g$) can be expressed as:
\begin{align}
    \omega_{g-insB} &= l_d(\big{[}t_{qv-g}, l_e(p)\big{]}) \\ &= l_d(\big{[}Q_g([\bm{t}_q, l_e([\phi, p])], v(i), p), l_e([\phi, p])\big{]}).
\end{align}
% \begin{subequations}
% \begin{align}
%   \bm{\omega}_q &= \ell(p, Q) = d(e(p, Q\big{(}v(i), p\big{)})), &&\text{[QFormer]} \\
%   \bm{\omega}_{gq} &= d(e(p), Q\big{(}v(i), p, e(p)\big{)}), &&\text{[Grounded]}
% \end{align}
% \end{subequations}

Note the two key differences from the standard QFormer \eqref{eq:1}, which is to use the encoded prompts as additional inputs to the QFormer. Second, QFormer output queries, along with the unaltered encoded prompts, are directly used to prompt language generation from the decoder. For better elucidation, we provide a visual comparison in Fig. \ref{fig:proposal}. 

 We note that this grounding directly informs the QFormer of the encoder latent space; meaning that the QFormer obtains direct feedback while modelling the distribution of the latent representations that is expected, and can be suitably decoded by the LLM decoder. However, conditioning only the decoder leads to a natural tradeoff: the potential ease of modeling may be offset by the reduced model capacity to incorporate said conditioning, since the encoder is no longer involved in interpreting the QFormer queries. Our experiments  show that benefits due to the former significantly outweigh the latter. Additionally, we find that splitting the LLM yields two other benefits: first, it reduces memory requirements for end-to-end training, as the encoder embeddings can now be precomputed. Second, inference/text generation is sped up significantly since decoding requires multiple forward passes through the entire LLM, whereas we only rely on the decoder.

% \begin{figure*}[!ht]
%      \centering
%          \includegraphics[width=0.8\linewidth]{figs/proposed_framework_updated_2.png}
%          \caption{Illustration of LLM grounding for QFormer, distinguishing it from InstructBLIP \cite{WenliangDLLTZWLFH2023}}.
%         \label{fig:proposal}
% \end{figure*}

\begin{table}
\begin{center}
\resizebox{\columnwidth}{!}{%
\begin{tabular} {|c |c |c | }
 \hline
 Model & Captioning (BLEU-4) & VQA (Acc.) (\%) \\
 \hline
 QFormer  & 0.238  & 57.72  \\
\hline
Grounded (Ours)  & $\bm{0.364}$  & $\bm{63.25}$  \\
\hline
\end{tabular}
}
\caption{Performance comparison for single task models; each model is only trained on one task (either Captioning or VQA) and evaluated on the corresponding validation set.}\label{tab:1}
\end{center}
\end{table}

\section{Experiments}\label{sec:experiment}
Here, we provide empirical evidence to support our hypothesis on more efficient training. Due to compute limitations, we restrict ourselves to smaller LLMs and datasets for our experiments. Specifically, we consider the FlanT5-base \cite{HyungCW2022} as our frozen LLM and the Eva-clip-g/14 \cite{QuanSFWWC2023}, a CLIP-based vision transformer, for the frozen image encoder. 

Thus, it is important to point out that we do not intend to compete against the state-of-art InstructBLIP models. Indeed, it is not possible to replicate anymore since the LAION-5B dataset, the key component of QFormer pretraining, was taken offline due to questionable content \cite{DavidT2023}. Furthermore, it would require billion parameter LLMs and large scale pretraining, which we exclude from the scope of this work. Rather, in our small but controlled setup, we aim to compare the efficiency of the default QFormer pipeline against ours. Finally, for both frameworks, the QFormer uses the default architecture and initialization as in \cite{JunnanLLSH2023, WenliangDLLTZWLFH2023}, without any prior pretraining. \\
\textbf{Tasks} For this study, we primarily focus on two tasks: Captioning and Visual Question Answering (VQA). For captioning we consider the COCO captions dataset \cite{XinleiCFLVGDZ2015} and for visual question answering we consider VQAv2 \cite{YashGKSBP2017} for training and OKVQA (Outside Knowledge VQA) \cite{KennethMRFM2019} for zero-shot evaluation. To specify the task, we fix a set of text prompts for each task during training and randomly sample one prompt from that set for each iteration.

\begin{table*}[!htb]
\begin{center}
\begin{tabular} {|c |c |c | c| }
 \hline
 Model & Pretrain Captioning (BLEU-4) & Final Captioning (BLEU-4) & VQA (Accuracy) (\%)  \\
 \hline
 QFormer  & 0.231  & 0.209 & 55.4  \\
\hline
Grounded (Ours)  & $\bm{0.357}$  & $\bm{0.362}$ & $\bm{66.8}$  \\
\hline
\end{tabular}
\caption{Performance comparison for pretraining, followed by multi-task instruction finetuning.}\label{tab:2}
\end{center}
\end{table*}

\subsection{Single Task Evaluation}
We begin evaluating our grounded QFormer against the default one, without any pretraining, to establish baselines on captioning and question answering. Both models are trained under multiple hyperparameter configurations until training loss saturation, which takes about 50 epochs, and the respective best validation scores are reported.  Our results in Table \ref{tab:1} indicate that our method significantly improves the learning capacity of the pipeline, even without any pretraining.

\subsection{Multi-task Training}
A commonly sought advantage of VLMs is their ability to simultaneously perform different tasks well. To achieve such a model, pretraining followed by multi-task training/finetuning is effective \cite{JunnanLLSH2023, WenliangDLLTZWLFH2023}. To emulate this, we train multi-task models to perform both Captioning and VQA, by first using image captioning as the pre-training objective (20 epochs), and then performing multi-task instruction finetuning for captioning and VQA (15 combined epochs). The choice of our datasets, Coco Captions and VQAv2---which share same train-val-test image split---allows us to prevent train-test information leak for a fair evaluation. Note that this procedure also mimics pretraining with captioning, followed by an instruction finetuning procedure for training BLIP  models. Results are given in Table \ref{tab:2}. We find that for the designated scale pretraining followed by instruction finetuning task, our method achieves significant improvements over the standard QFormer. Additionally, we see that our method beats the single task baseline for VQAv2 (Table~\ref{tab:1}), harnessing the benefits of multi-task training and the synergy between captioning and VQA tasks.

\subsection{Zero-shot evaluation}

Instruction finetuned VLMs can demontrate remarkable zero-shot abilities and in the same vein, we evaluate the models on OKVQA, with results in Table~\ref{tab:3}. For comparison, we report the zero-shot OKVQA performance for BLIP-2 models \cite{JunnanLLSH2023} (InstructBLIP includes OKVQA during training, so comparison unavailable). Surprisingly, we find that our method demonstrates comparable zero-shot performance, despite utilizing two orders of magnitude less pretraining and much smaller LLMs.

\begin{table}[!htb]
\begin{center}
\resizebox{\columnwidth}{!}{\begin{tabular} {| c | c | c | }
 \hline
 Model & OKVQA (Acc.) (\%) & LLM \\
 \hline
 QFormer (Baseline) & 28.83 & FlanT5-base  $\sim$ 240M \\
\hline
Grounded (Ours)  & $\bm{38.96}$ &  FlanT5-base $\sim$ 240M\\
\hline
 BLIP-2 OPT  & $36.4$  & OPT $\sim$ 6.7B\\
\hline
 BLIP-2
FlanT5-xl  & {40.7}  & FlanT5-xl $\sim$ 3B\\
\hline
\end{tabular}}
\caption{Zero-shot performance comparison on OKVQA.}\label{tab:3}
\end{center}
\end{table}

\subsection{Pretraining Efficiency}

\begin{figure}
     \centering
         \includegraphics[width=\linewidth]{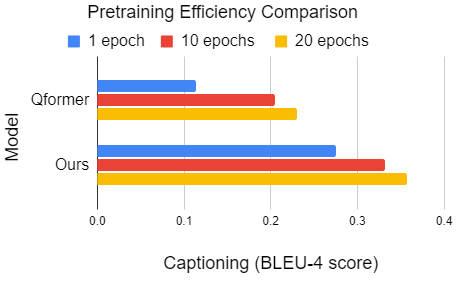}
         \caption{Pretraining comparison, showcasing our framework significantly improves pretraining efficiency.}.
        \label{fig:pretrain}
\end{figure}

Next, we compare the pretraining efficiency of both pipelines. Specifically, we evaluate the best validation score achieved for the captioning task within a fixed number of epochs, across different hyperparameter and learning rate configurations. Across all runs, we report the highest performance achieved within a fixed number of epochs. As shown in Fig.~\ref{fig:pretrain}, we find that our method achieves significantly faster learning and better peak performance. Moreover, our modification to the original architecture allows the Qformer output to be integrated into an intermediate layer of the language model rather than the initial layer. This adjustment reduces computational overhead in the earlier layers, resulting in a training speed-up per epoch of 2. 1\%, as shown in Table ~\ref{tab:4}. While this efficiency gain seems modest, it is due to our current reliance on the language encoder to provide inputs to the Qformer. However, in more recent VLM architectures with decoder-only LLMs, such as variants of LLaVA ~\cite{liu2024visual, liu2024improved}, this computation is avoided. Consequently, we anticipate more significant training speedups in these newer architectures.

\begin{table}
\begin{center}
\resizebox{0.4\columnwidth}{!}{%
\begin{tabular} {|c |c |}
 \hline
 Qformer &  Ours\\
 \hline
 1.00  & \textbf{0.979}   \\
\hline
\end{tabular}
}
\caption{Comparison of training times per epoch on the captioning dataset between the Qformer-based model and our model. The values are normalized relative to the Qformer-based model}\label{tab:4}
\end{center}
\end{table}

\subsection{Grounding Ablation}

\begin{figure}
     \centering
         \includegraphics[width=\linewidth]{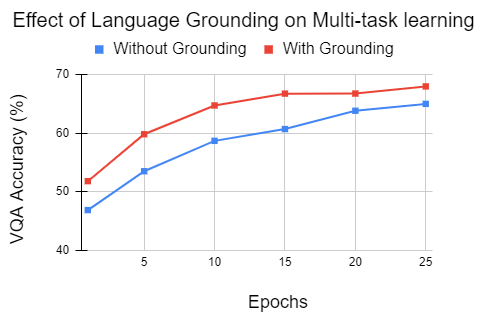}
         \caption{Grounding the QFormer with LLM encoder representations can significantly speed up multi-task learning, especially when the prompts are more informative of the task (as in VQA).}.
        \label{fig:grounding_ablate}
\end{figure}

Finally, we study the effect of grounding i.e. feeding the encoded prompts to the QFormer. With the multi-task setup, we first pretrain two models, one with grounding and one without, to roughly the same captioning score, noting that the model without grounding takes an additional five epochs to reach similar performance. However, subsequent multitask instruction fine-tuning reveals more pronounced advantages. Specifically, while the captioning score remains similar, we find that language grounding significantly speeds up VQA learning, with the grounded model maintaining a $5$--$6$\% performance lead in the earlier epochs and hitting peak performance much faster (see Fig.~\ref{fig:grounding_ablate}). This jives with our intuition that grounding aids the QFormer in learning the appropriate encoder latent representations for prompting the decoder, while noting that this speed-up is more significant for VQA, which incidentally generates more informative input prompts, i.e., image-based questions in this case.

\section{Towards a General Framework}

Our experiments in the previous section show promise in terms of improving the efficiency of the standard QFormer pipeline. However, there are certain caveats. First, our experiments relied on encoder-decoder LLM architectures, whereas most modern state-of-the-art models rely on decoder-only architectures \cite{touvron2023llamav2, openai2024gpt4}. Thus, although there is evidence to support decoder-only representations imbue semantics \cite{AlecRKHKS2021}, it is unclear a) whether it can align with QFormer latents and b) what choice of intermediate layer representations is suitable for being modelled by the QFormer. Second, note that in our experiments, the QFormer is in its native BERT initialization. However, it is possible that after pretraining Stage 1, which involves joint vision-language training, the latents of the QFormer are affected such that our hypothesis about modelling the semantics vs. syntactics no longer holds. Even so, since the Stage 1 trained QFormer serves as a universal plug-and-play model, we would still like to utilize it when scaling up our framework. Therefore, we need to verify whether our hypothesis still holds. 

\begin{figure}
     \centering 
         \includegraphics[width=0.95\linewidth]{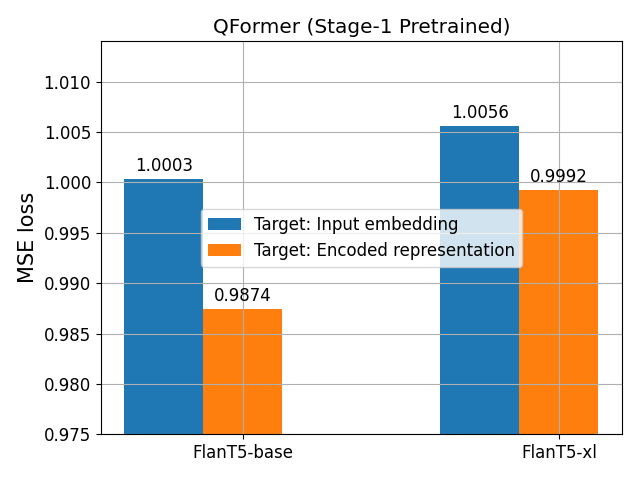}
         \caption{Experiment to study ease of modeling input embeddings vs encoder representations, the lower error indicates that the QFormer (Stage-1 pretrained) finds it easier to model the latter.}
        \label{fig:flan_pretrained}
\end{figure}

For answering these questions comprehensively, we leave the more compute-intensive experiments for future work. However, we offer some preliminary evidence by relying on our regression setup as before. Specifically, for the encoder-decoder models, we repeat the same experiment, but with the QFormer weights initialized from the pretraining Stage 1 checkpoint weights to check whether it changes the error trend. Fig. \ref{fig:flan_pretrained} again shows that even after pretraining, the QFormer representations are transferred to the encoder representations with more ease.
% add plot

\begin{figure}
     \centering 
         \includegraphics[width=\linewidth]{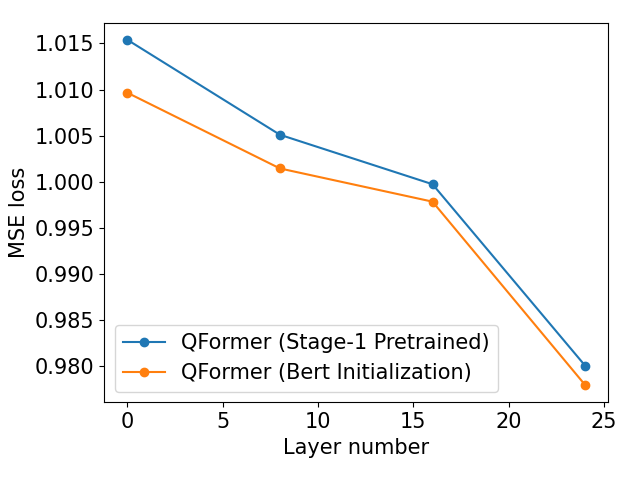}
         \caption{Experiment to study ease of modeling outputs from different layers of decoder-only LLM, open-llama-3b-v2. The results indicate that intermediate representations from deeper layers are easier to model, lending credence to our semantics hypothesis.}
        \label{fig:llama_3b}
\end{figure}

Next, to test our hypothesis for the decoder-only models, we consider the open-llama-3b-v2 \cite{XinyangGL2023, together2023redpajama}, an open source reproduction of llama-3b-v2 \cite{touvron2023llamav2}, consisting of twenty-four layers before the final decoding layer. To compare different intermediates, we consider a representative set of layers $\{0, 8, 16, 24\}$, where $0$ refers to the input embeddings and for every other layer $n$, the regression targets are the outputs of the $n^{th}$ layer. Since LLM outputs across layers can vary significantly in norm, we first normalize each target distribution to standard zero-mean and unit variance. The QFormer outputs are then projected using a linear layer to model the corresponding LLM representations. The respective errors are compared in Fig. \ref{fig:llama_3b}, and we find that it is increasingly easier for the QFormer to model the LLM intermediates, the deeper we go into the network. Similar trends of increasing predictive power and clustering performance across layers within decoder-only models are observed by \cite{wu2023meta, wu2024transformer}.

% add plot

\section{Discussion \& Future Work}
In this work, we presented an alternative pipeline for QFormer-based vision-language fusion. We first showed that the QFormer latents are better suited to modelling the LLM intermediate outputs. Using this insight, we proposed our modified framework, which uses the QFormer to directly condition generation in the LLM latent space, while also augmenting the pipeline with grounding for ease of latent modeling. Our experiments with encoder-decoder LLMs demonstrated an improvement in both performance and learning efficiency, when compared to the standard QFormer pipeline. Finally, we also highlighted potential for extending our method to decoder-only models. 

However, there remains a performance gap to state-of-the-art models, arising from the limited LLM capacity and pretraining scale. Indeed, in our preliminary experiments with larger models, we observed performance dips for both methods when pretrained only on COCO captions ($\sim$ 500k samples), indicating a need for scaling up pretraining, potentially to InstructBLIP levels ($\sim$ 130m samples).  A second limitation lies in the fact that extending our framework to decoder-only models may not be trivial; even though representation modelling may be easier in the intermediate layer, we also need enough depth in the model to be able to assimilate the visual information fed via the QFormer queries. We plan to explore this trade-off in future work by scaling up to: a) more pretraining and b) larger LLMs for better generative capabilities.  

\clearpage
%\bibliographystyle{IEEEbib}
%\bibliography{refs.bib}

{
    \small
    \bibliographystyle{ieeenat_fullname}
    \bibliography{main}
}


\end{document}

%% file: preamble.tex
\usepackage[dvipsnames]{xcolor}